# An Optimized Architecture for Unpaired Image-to-Image Translation

Mohan Nikam

Medi-Caps Institute of Science and Technology, Indore
mohannikam19@gmail.com

**Abstract.** Unpaired Image-to-Image translation aims to convert the image from one domain (input domain A) to another domain (target domain B), without providing paired examples for the training. The state-of-the-art, Cycle-GAN demonstrated the power of Generative Adversarial Networks with Cycle-Consistency Loss. While its results are promising, there is scope for optimization in the training process. This paper introduces a new neural network architecture, which only learns the translation from domain A to B and eliminates the need for reverse mapping (B to A), by introducing a new Deviation-loss term. Furthermore, few other improvements to the Cycle-GAN are found and utilized in this new architecture, contributing to significantly lesser training duration.

## 1 Introduction

Transferring characteristics from one image to another is at the core of image-to-image translation. The data is said to be unpaired, when there is no one-to-one correspondence between training images from input domain A and the target domain B. For example, converting an image of apple to an image of orange or image of a horse into an image of a zebra, etc. The challenge is also to keep the background of the image intact. Everything except the apple (in case of conversion from apple to orange) must not get altered. Therefore, unpaired image-to-image translation becomes an even difficult problem than having paired examples. However, there are many scenarios where such translation is necessary. Also, collecting unpaired data is much easier than deriving a paired set of images. The paper "Unpaired Image-to-Image Translation using Cycle-Consistent Adversarial Networks", famously known as Cycle-GAN by Zhu et al [1] offers an approach for creating such a system. In order drive the learning process for creating such a system, they introduced Cycle-consistency loss, which involves not only learning the mapping for conversion A to B, but also the transitive mapping (B to A). This works very well for unpaired translation in many kinds of images. However, the learning of reverse mapping turns out to be a redundant step which takes equal amount of computational time to train, as it is consumed for training the translation A to B.

This paper proposes a new architecture, which learns only the forward mapping (from domain A to B) and eliminates the need of backward mapping (from domain B to A) while keeping the goodness achieved by this backward mapping (directing the training to proceed in right direction), by introducing a new loss term, *Deviation-Loss*. The training duration for this architecture is found to achieve a significant speedup as compared with Cycle-GAN.

## 2 Related Work

**Cycle-GAN.** Unpaired Image-to-Image Translation Cycle-Consistent Adversarial Networks by Zhu et al. [1] provided an approach for this problem using Cycle-Consistency loss. Further comparison with their approach is presented in next section.

**Generative Adversarial Network (GAN).** This paper by Goodfellow et al. [2] involves a generator and a discriminator optimizing against each other in a 2-player zero-sum game. Generator tries to fool the discriminator and discriminator tries to correctly classify a real image and a fake image, coming from the output of generator.

**AutoEncoders.** It involves 2 neural networks, one is used to encode an image and another is used to decode it. Purpose of using AutoEncoders [3] is for dimensionality reduction or simply, compression of images. In this paper, Convolutional AutoEncoder is used, sandwiched between a translator (discussed in next section). In Convolutional AutoEncoder, encoder consists of convolutional neural network [4, 12, 13], instead of fully connected neural network, which is more effective for encoding image. Likewise, for decoding the image, Deconvolution is performed, to bring the same image from the encoded image.

**Neural Artistic Style Transfer.** This work [6] and its successors [10] [11] deals with transferring artistic characteristics from one image (preferably painting) to another image (a photograph). For doing so, they proposed extracting style of style-image by taking feature encodings from the intermediate layers of Convolutional Neural Network, Pretrained to encode the features effectively and obtaining its gram-matrix. And combining this extracted style with the extracted content, by getting the feature encodings from lower layers of the same convolutional neural network.

**Disco-GAN.** This work [7] is similar to Cycle-GAN for unpaired Image translation, with difference that their approach involves two reconstruction losses instead of one single cycle-consistency loss and different measures for distance between the images. However, their approach too involves learning the reverse mapping (B to A).

**Conditional-GAN.** Image-to-Image translation with Conditional Adversarial Networks by Isola et al. [8] have proposed an approach for image-to-image translation, but by using paired examples for the training process. Their approach involved using Generative Adversarial Networks [2] in conditional setting.

# 3 Proposed Approach

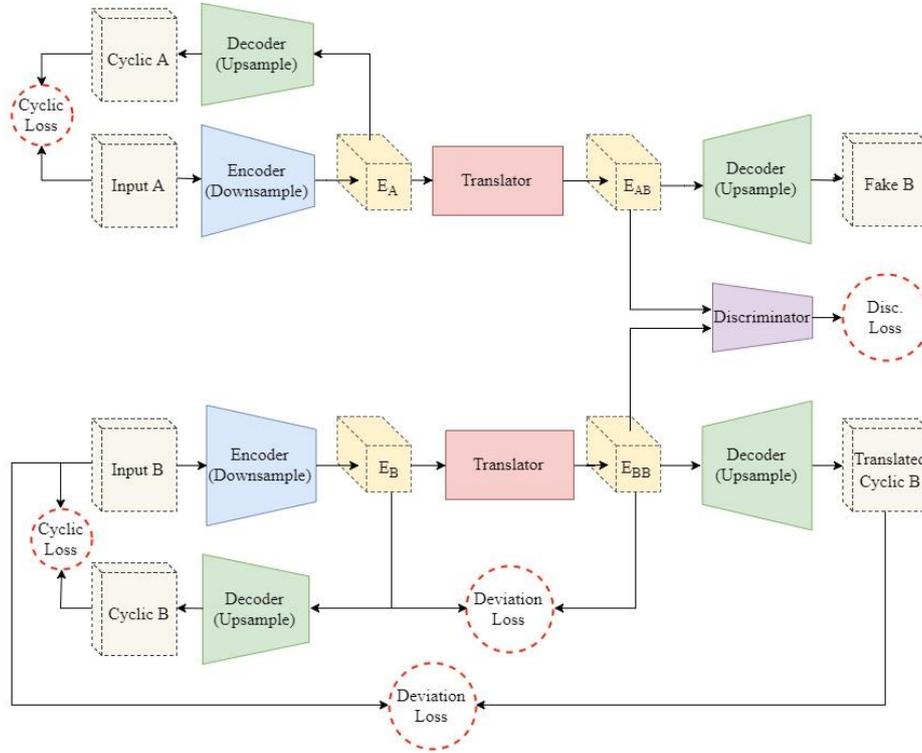

**Figure 1.** Proposed Architecture. The aim is to obtain image *Fake B*, from the image *Input A*. Neural networks are represented by solid boundaries and those having same color represents same network. The orange colored circles indicate *loss terms*.

There are 4 different Neural Networks, each would perform a specific function after the completion of training process as follows:

i. *Encoder:* This network encodes the input image from any domain (A or B), by performing convolution [4, 12, 13] operations, to down-sample the image. This is different from Encoder in Cycle-GAN, which used separate encoders for A and B.

ii. *Decoder:* This network learns to decode the encoded image from any domain (A or B), by performing deconvolution operation, to up-sample the encoded image. This is different from Cycle-GAN as well, since it used separate decoders for A and B.

iii. *Translator:* It is created using Residual Networks [5]. It inputs the encoded features from encoder (of both domains A or B) and ensures that output is only encoding from domain B. i.e., it selectively translates the encodings from domain A, while allows images from domain B, to be pass unmodified. Thus, only one Translator is used here whereas in Cycle-GAN, there were 2 translators (they call it Generators A2B and B2A).

iv. *Discriminator:* This network inputs the encoded image and identifies if the encoding is real or fake. This is different from Cycle-GAN, as its discriminator takes in the whole image and not its encodings.

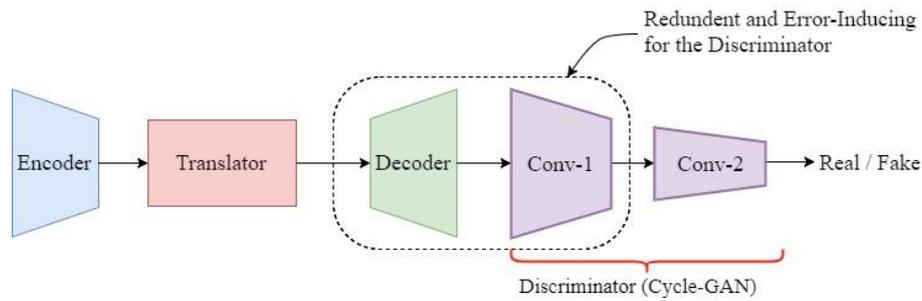

**Figure 2.** Cycle-GAN up-samples the translated encoding and discriminator needs to again down-sample it to the same level*(Conv-1)* and further convolute*(Conv-2)* for classification.

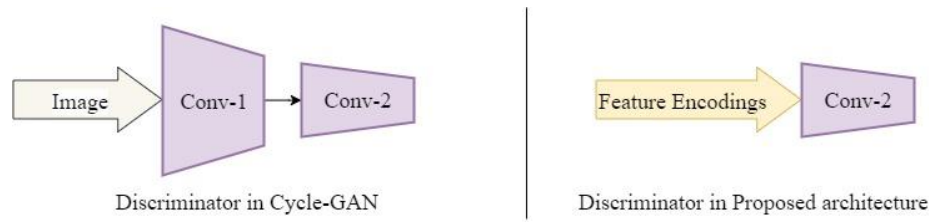

**Figure 3.** Cycle-GAN Discriminator vs. Discriminator in proposed architecture, which takes down-sampled encodings as input as opposed to whole image. This results in lesser complex discriminator and still not compromising in the accuracy.

Following are the loss functions used to train and optimize the above mentioned neural networks:
i. *Cyclic Loss:*
This loss is for encoder and decoder network. It works the same way as for AutoEncoders. The loss term is defined as the difference between the input image and the cyclic image (which is obtained by encoding and then decoding back the input image).
ii. *Deviation Loss:*
This loss is for the Translator Network. It is defined as the sum of following 2 terms:
a. Difference between the encodings of image from domain B, before and after passing through the translator.
b. Difference between the input image from domain B and the image obtained by encoding, translating and then decoding the image (translated cyclic B).[1]

This loss is introduced as a replacement for the cycle-consistency loss which was present in the Cycle-GAN architecture.

---

[1] This term is yet to be tested for importance. However, it does not hurt the performance.

The Deviation Loss regularizes the training of Translator, by directing the translator to translate only the bare-minimum part of encoded image of domain A, to make it appear like real encoding from domain B. Also, it enforces the spatial features to be kept intact throughout the translation.

iii. *Discriminator Losses:*

These losses are for optimization of Translator network and the Discriminator Network. If the Discriminator correctly classifies the encodings of the fake image, the Translator Network is penalized. If the Discriminator incorrectly classifies either of real or fake encodings, the Discriminator network is penalized. This loss is kept similar to that of Cycle-GAN, but the structure of Discriminator has changed in the proposed architecture.

**Explanation.** Considering an example for converting the image of an apple to an orange. The goal is to perform this task while keeping the background intact. Forward pass involves downsampling of the input image of an apple, translating it to the encoding of orange and upsampling it, to produce the image of an orange. Deviation loss ensures the output of translator is always the feature encodings of orange. Thus, the image of orange is unchanged (including background), whereas the image of apple changes in such a way that apples are converted into oranges (since discriminator network is forcing this conversion) while everything in the background is unchanged (since deviation-loss is resisting this change). The key idea is that translator network leans not to alter the background and orange but to necessarily convert the apple to orange.

## 4  Experimental Evaluation

The performance of this architecture is compared with the Cycle-GAN implementation, on Tensorflow Framework, on Intel® AI DevCloud using Intel Xeon Gold 6128 processors.

**Table 1.** Comparison of time taken by Cycle-GAN and proposed architecture.

| No. of Epoch(s) | Time by Cycle-GAN | Time by Proposed Architecture | Speedup |
|---|---|---|---|
| 1 | 66.27 minutes | 32.92 minutes | 2.0128x |
| 2 | 132.54 minutes | 65.84 minutes | 2.0130x |
| 3 | 198.81 minutes | 98.76 minutes | 2.0138x |
| 15 | 994.09 minutes | 493.80 minutes | 2.0131x |

Furthermore, it is observed that due to using same neural network for encoding and decoding, and also using less complex Decoder, the proposed system converges nearly twice as fast i.e., it needs nearly half the number of epochs required by Cycle-GAN to produce the same result.

Thus, net effective speedup could reach 4 times the Cycle-GAN.

The work on this is still ongoing to present wider variety of results.

## 5   Result and Conclusion

The neural networks were trained on images of apple and oranges collected from ImageNet and were directly available from [9]. The images were of the dimension 256x256 pixels. The training set consists of 1177 images of class apple and 996 images of class orange.

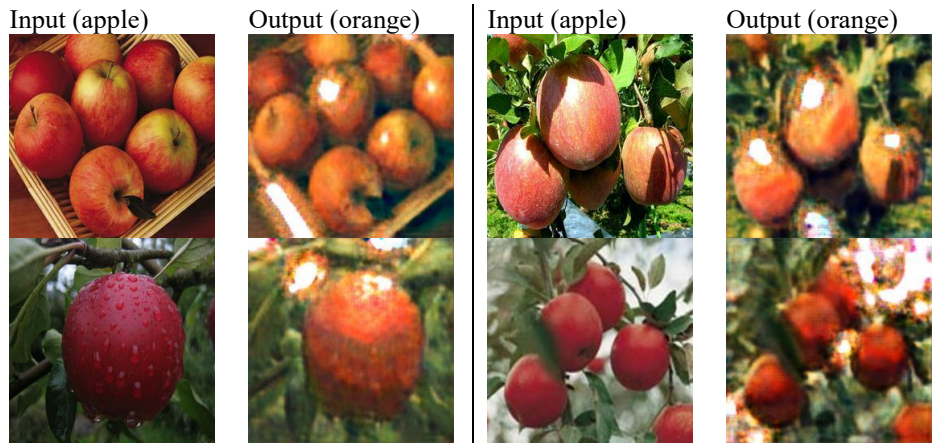

Input (apple)   Output (orange)   Input (apple)   Output (orange)

The important aspects of the proposed architecture are:
1. Elimination of second translator (to translate B to A)
2. Using same Neural Network to encode images from both domains (A or B), and same Neural Network to decode images from both domains (A or B).
3. Discriminator takes down-sampled image encoding as input, as opposed to taking whole image which was the case with discriminator in Cycle-GAN.
4. Use of Deviation Loss, in lieu of Cycle-Consistency loss, from Cycle-GAN.

## 6   Conclusion

This paper introduces a new architecture for Unpaired Image-to-Image Translation.

The Proposed architecture is able to simplify the architecture of state-of-the-art, Cycle-GAN and hereby achieving experimental speedup of 2.01x for the training process, which under subjective consideration for convergence can be as much as 4x.

**Acknowledgments.** I would like to give my sincere thanks to Prof. Dheeraj Rane, HOD, CSE, Medi-Caps University, for his invaluable support and constant encouragement. The computational resources for this work were provided by Intel Corporation on Intel AI DevCloud under the Student Ambassador for AI program. The work was also partly supported by Amazon for AWS EC2 instances through their student scholarship.

# References

1. T. P. P. I. A. A. E. Jun-Yan Zhu, "Unpaired Image-to-Image Translation using Cycle-Consistent Adversarial Networks," [Online]. Available: https://arxiv.org/abs/1703.10593.
2. J. P.-A. M. M. B. X. D. W.-F. S. O. A. C. Y. B. Ian J. Goodfellow, "Generative Adversarial Networks," [Online]. Available: https://arxiv.org/abs/1406.2661.
3. Autoencoders – Chapter 14, The deep Learning Book, Ian J. Goodfellow. Available: http://www.deeplearningbook.org/contents/autoencoders.html.
4. A. Z. Karen Simonyan, "Very Deep Convolutional Networks for Large-Scale Image Recognition," [Online]. Available: https://arxiv.org/abs/1409.1556.
5. X. Z. S. R. J. S. Kaiming He, "Deep Residual Learning for Image Recognition," [Online]. Available: https://arxiv.org/pdf/1512.03385.pdf.
6. A. S. E. a. M. B. L. A. Gatys, "A Neural Algorithm of Artistic Style. arXiv:1508.06576v2 [cs.CV]," [Online]. Available: https://arxiv.org/abs/1508.06576v2.
7. M. C. H. K. J. K. L. J. K. Taeksoo Kim, "Learning to Discover Cross-Domain Relations with Generative Adversarial Networks" [Online]. Available: https://arxiv.org/abs/1703.05192.
8. J. Y. Z. T. Z. A. A. E. Phillip Isola, "Image-to-Image Translation with Conditional Adversarial Networks" [Online]. Available: https://arxiv.org/abs/1611.07004.
9. T. Park, UC Berkeley, apple2orange Dataset, Cycle-GAN. Available: people.eecs.berkeley.edu/~taesung_park/CycleGAN/datasets/apple2orange.zip
10. M. B. A. H. E. S. Leon A. Gatys, "Preserving Color in Neural Artistic Style Transfer," [Online]. Available: https://arxiv.org/abs/1606.05897.
11. Y. N. Roman Novak, "Improving the Neural Algorithm of Artistic Style," [Online]. Available: https://arxiv.org/abs/1605.04603.
12. I. S. a. G. H. A. Krizhevsky, "Imagenet classification with deep convolutional neural networks," NIPS, 2012.
13. K. Fukushima, "Neocognitron: A Self-organizing Neural Network Model for a Mechanism of Pattern Recognition Unaffected by Shift in Position," Biological Cybernetics, 1980.